\documentclass{article}

\usepackage{arxiv}

\usepackage[utf8]{inputenc} 
\usepackage[T1]{fontenc}    
\usepackage{hyperref}       
\usepackage{url}            
\usepackage{booktabs}       
\usepackage{amsfonts}       
\usepackage{nicefrac}       
\usepackage{microtype}      
\usepackage{lipsum}
\usepackage{graphicx}
\usepackage{subcaption}
\usepackage{geometry}
\usepackage{capt-of}
\usepackage{tabularx}
\usepackage{graphicx,subcaption}
\usepackage{calc}
\usepackage{amsmath}
\graphicspath{ {./images/} }

\title{Hand gesture detection in tests performed by older adults}

\author{
 Guan Huang \\
  School of Information and Technology\\
  University of Tasmania\\
  Sandy Bay, TAS 7005 Australia\\
  \texttt{guanh@utas.edu.au} \\
   \And
 Son N. Tran \\
  School of Information and Technology\\
  University of Tasmania\\
  Sandy Bay, TAS 7005 Australia\\
  \texttt{son.tran@utas.edu.au} \\
  \And
 Quan Bai \\
  School of Information and Technology\\
  University of Tasmania\\
  Sandy Bay, TAS 7005 Australia\\
  \texttt{quan.bai@utas.edu.au} \\
  \And
 Jane Alty \\
  Wicking Dementia Research and Education Centre\\
  University of Tasmania\\
  Hobart, TAS 7000 Australia \\
  \texttt{jane.alty@utas.edu.au} \\
}

\begin{document}
\maketitle
\begin{abstract}
Our team are developing a new online test that analyses hand movement features associated with ageing that can be completed remotely from the research centre. To obtain hand movement features, participants will be asked to perform a variety of hand gestures using their own computer cameras. However, it is challenging to collect high quality hand movement video data, especially for older participants, many of whom  have no IT background. During the data collection process, one of the key steps is to detect whether the participants are following the test instructions correctly and also to detect similar gestures from different devices. Furthermore, we need this process to be automated and accurate as we expect many thousands of participants to complete the test. We have implemented a hand gesture detector to detect the gestures in the hand movement tests and our detection mAP is 0.782 which is better than the state-of-the-art. In this research, we have processed 20,000 images collected from hand movement tests and labelled 6,450 images to detect different hand gestures in the hand movement tests. This paper has the following three contributions. Firstly, we compared and analysed the performance of different network structures for hand gesture detection. Secondly, we have made many attempts to improve the accuracy of the model and have succeeded in improving the classification accuracy for similar gestures by implementing attention layers. Thirdly, we have created two datasets and included 20 percent of blurred images in the dataset to investigate how different network structures were impacted by noisy data, our experiments have also shown our network has better performance on the noisy dataset.
\end{abstract}


\section{Introduction}
There is increasing interest in developing computer tests that can completed remotely, away from the research centre. This provides convenience for participants, facilitates involvement in research for participants who live remotely and allows research studies to continue when restrictions on travel occur, such as during the COIVD 2019 pandemic. Our team are developing a new online test that analyses hand movement features associated with ageing that can be completed remotely from the research centre. Our test has included two different tests, the alternating hand movement test involves opening and closing the whole hand and the alternating finger-tapping test involves opening and closing the fingers of each hand repeatedly. Both tests are considered an important test for evaluating motor function. Figure 1 shows how the finger-tapping test is performed, the participants will be instructed to switch between ‘Gesture 1’ and ‘Gesture 2’ quickly and repeatedly.
Since the video tests will request participants to open and close their fingers as fast as possible using relatively low fps laptop cameras, motion blur may be a significant problem during the data collection process. Meanwhile, getting the participants to follow the instruction and recording the correct gestures are essential for our study as most participants will complete the tests unsupervised in their own homes. To obtain high-quality video data in the hand movement tests, it is important to detect if users are following the instructions with the correct gestures.
\begin{figure}[h]
    \centering
        \begin{subfigure}[b]{.4\linewidth}
        \centering\includegraphics[width=.8\linewidth]{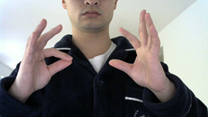}
        \caption{Gesture 1}
        \end{subfigure}
        \begin{subfigure}[b]{.4\linewidth}
        \centering\includegraphics[width=.8\linewidth]{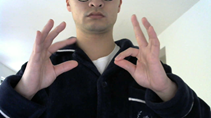}
        \caption{Gesture 2}
        \end{subfigure}
    \caption{Finger tapping test demonstration}
    \label{fig:my_label1}
\end{figure}
\newline
Since the hand movement tests will request participants to open and close their fingers as fast as possible using relatively low fps laptop cameras, motion blur may be a significant problem during the data collection process. Meanwhile, getting the participants to follow the instruction and recording the correct gestures are essential for the fol-lowing study as most participants will complete the tests unsupervised in their own homes. To obtain high-quality video data in the hand movement tests, it is important to detect if users are following the instructions with the correct gestures. In this case, hand gesture recognition will apply important implications in guiding participants to assist data collection for TAS Test hand movement tests.
This paper has the following three contributions:
\begin{enumerate}
    \item We have processed and labelled 7071 images, then built a hand gestures classification dataset for dementia-related gesture classification and motion blur study
    \item We have compared the performance of different network structures on the hand gesture classification task  
    \item We proposed a novel approach and used the attention technique to increase the classification performance on similar gestures. Our model is better than state-of-the-art.
    \item We have demonstrated how the quality of images will impact the performance on hand gesture classification through two sets of experiments and have proven our net-work structure has better result than state-of-the-art on noisy dataset.
\end{enumerate}

\section{Related Work}
\subsection{Hand gesture detection}
The history of hand pose detection is originated from 1980s, data gloves and vision based methods are regarded as the most common methods for hand pose detection. Data gloves are wearable devices with sensors attached to the fingers or joints of the glove, then the researchers use the tactile switches, optical goniometer or resistance sensors to measure the bending of finger joints to estimate the hand poses \cite{r4}. This method mainly focus on increase the accuracy to pinpoint the position of the hand in the 3d model. According to Ahmed et al.\cite{r5}, commercial data gloves are normally less affordable for large adoption because they are priced  between  \$1000  and  \$20,000. 
\newline
With the development of the artificial intelligence and deep learning \cite{lecun2015deep}, vision based methods have evolved dramatically and deep learning based method is now dominated the hand pose detection field. Table 1 shows different deep learning based methods solving hand gestures detection problem.

    \begin{table}[ht]
    \centering
    \resizebox{\textwidth}{!}{\begin{tabular}{p{0.20\textwidth} p{0.20\textwidth} p{0.31\textwidth} p{0.10\textwidth} p{0.08\textwidth} p{0.08\textwidth} p{0.08\textwidth}}
    \hline
        Task & Model & Gestures & Gestures Number & Accuracy & Year & Paper \\
    \hline
        Classification & 3DCNN model & Arabic sign Language & 40 & 90.79 & 2020 &  \cite{al2020hand} \\

        Classification & ConvLSTM model &  Dynamic Hand Gesture & 14 & 96.07 & 2020 &  \cite{do2020robust} \\

        Classification & EDenseNet & American Sign Language & 26 & 98.50 & 2021 &  \cite{tan2021hand} \\
    
        Object Detection & ResNet-10 & nvGesture Dataset & 25 & 77.39 & 2019 &  \cite{kopuklu2019real} \\

        Object Detection & Mobilenets-SSD & Custom Dataset & 4 & 93.01 & 2020 &  \cite{liao2021occlusion} \\

        Object Detection & SSD & American Manual Alphabet & 10 & 93.60 & 2018 &  \cite{yi2018long} \\
    
        Object Detection & YOLOV3 & Custom Dataset & 5 & 97.68 & 2021 &  \cite{mujahid2021real} \\
        
    \hline
    \end{tabular}}
    \caption{Table of Deep Learning based methods for hand gesture detection}
    \label{tab:table_deep2}
\end{table}
\subsection{Object detection algorithm}
Object detection is often regarded as the fundamental to many computer vision tasks. Object detection not only focuses on detecting visual objects such as vehicles, humans, dogs, and many other objects, but also addresses one of the most important questions: "What Objects are where?" \cite{zou2019object}. In recent years, the object detection methods can be categorised into two groups, two-stage detectors and one-stage detectors. Two-stage detectors are good at improving the detection and one-stage detectors are known as having faster detection speeds.
\newline
R-CNN algorithm family is an example of one-stage detectors, which includes R-CNN, Fast RCNN, and Faster RCNN \cite{girshick2015fast}\cite{girshick2014rich}\cite{ren2015faster}. The basic process for R-CNN algorithm is to propose several  regions that are likely to be objects, features are extracted on these candidate regions (using CNN) for classification. The object detection tasks are divided into two parts in the R-CNN algorithm, classification task and regression task.
\newline
YOLO is first introduced by Redmon in 2016. The input image will be divided into an S x S grid, and the YOLO algorithm treats the detection tasks as one regression problem, each grid cell will generate a predict bounding box and its prediction probability. This type of algorithms has largely increased the speed of detection whereas literature shows they lack capabilities in detecting small objects \cite{redmon2016you}. The YOLO object detection model has updated and improved several times. From first version to YOLOV5, there are many techniques implemented to improve the performance, including batch normalisation, anchor boxes, multi-scale training, feature pyramid networks for object detection, mosaic data augmentation and model size reduction etc \cite{redmon2017yolo9000}\cite{redmon2018yolov3}\cite{bochkovskiy2020yolov4}\cite{glenn_jocher_2021_4679653}.
\subsection{Feature Extraction Network}
The idea of convolutional neural network is first proposed by Hubel and Wiesel in the year of 1968, they found that each of human individual neuron does not respond to recognise the whole image, different neuron is only interested in specific inputs and will automatically ignore other information, for example, some neurons are only interested in straight horizontal lines while some are only interested in vertical lines \cite{hubel1968receptive}. Based on their study, LeCun et al. successfully implemented the idea of CNN in the computer vision field for image classification, their network Le-net-5 is also known as the prototype of the modern convolutional neural networks \cite{lecun1998gradient}.
\newline
Prior the implementation of deep learning, traditional image classification methods mainly used machine learning algorithms, however, such method performs badly with the increasing variability and resolution of the images.There are several important contributions made by AlexNet for solving the image classification problem. Firstly, AlexNet used ReLU activation function to replace traditional sigmoid function, this solved the problem of Gradient Vanishment. Secondly, they used dropout to disable a portion of neurons in the network to avoid over-fitting problems. Thirdly, they used max pooling instead of average pooling to avoid blur effect. Moreover, they proposed Local Response Normalisation Layer to avoid over-fitting. Lastly, they have proved training on multi-GPU had largely increased the training speed and process \cite{NIPS2012_c399862d}.
\newline
A significant number of modern backbone networks are derived from Resnet. Resent introduced a short-cut connection and set the gradient of the identity is 1, so that when passing the deep gradient into the bot-tom layer, the gradient is guaranteed not to disappear, which solved the problem of gradient vanishment. Another contribution for deep neural networks is to realise learning nothing can also be important. This module allows the model to choose for itself whether to update the parameters or not \cite{he2016deep}.
\newline
EfficientNet is another classic work in the image classification area, it summarised some approaches for model scaling, for example by increasing the channel size of the model to make it wider, and then by increasing the number of layers of the model to make it deeper. The main contribution of this work is to propose a way to scale down or up the model to fit the computational ability for different tasks \cite{tan2019efficientnet}.
\newline
Vision Transformer (ViT) has implemented transformer in the field of computer vision. The implementation of transformer enables the neural network to learn the position information of the images and increase the classification ability \cite{dosovitskiy2020image}. 
\section{Dataset}
\subsection{Our Dataset – A new dataset for hand gesture detection}
The dataset used in this research will be a private dataset TAS Test hand movement dataset. The dataset contains more than 2,500 participants' test videos from previous experiments. The participants are in-between age from 50 years old to 90 years old. The data is collected from different web cameras. In this dataset, we have used ‘ffmpeg’ to extract more than 20,000 images from TAS Test videos. Then we selected 5996 clear images and 1075 blurred images from TAS Test hand movement video tests. This dataset is not only useful for hand gesture recognition related to ageing but also include 20 percent of blurred image for the analysis of motion blur study. The detailed information of our dataset is also listed in the Table 2.
    \begin{table}[ht]
    \centering
    {\begin{tabular}{p{0.18\textwidth} p{0.12\textwidth} p{0.10\textwidth} p{0.15\textwidth} p{0.08\textwidth} p{0.08\textwidth} p{0.08\textwidth}  p{0.12\textwidth}}
    \hline
        Datasets & Total Images & Population & Hand Side & Left-Right & Age-Recorded & Include motion-blur \\
    \hline
        11k Hands \cite{afifi201911kHands} & 11,076 & 190 & Palm-dorsal & Both & Yes & No \\
        
        CASIA \cite{sunordinal} & 5,502 & 312 & Palm & Both & No & No \\
        
        Bosphorus \cite{inproceedings} & 4,846 & 642 & Palm & Both & No & No \\
        
        llTD \cite{kumar2008incorporating} & 2,601 & 230 & Palm & Both & No & No \\
        
        GPDS150hand \cite{ferrer2007low} & 1,500 & 150 & Palm & Right & No & No \\
        
        TASTest(Ours) & 7,071 & 2,500 & Palm-dorsal & Both & Yes & Yes \\
        
    \hline
    \end{tabular}}
    \caption{Table of datasets for hand gestures detection comparison}
    \label{tab:table3}
\end{table}
\subsection{Developing the datasets}
There are 5 different hand gestures included for the hand gesture classification tasks, they are ‘Open’, ‘Close’, ‘Pinch Open’, ‘Pinch Close’, and ‘Flip’ as shown in Figure 2.
\begin{figure}[h]
    \centering
        \begin{subfigure}[b]{.3\linewidth}
        \centering\includegraphics[width=.5\linewidth]{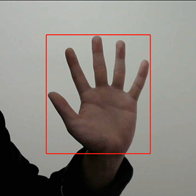}
        \caption{Open}
        \end{subfigure}
        \begin{subfigure}[b]{.3\linewidth}
        \centering\includegraphics[width=.5\linewidth]{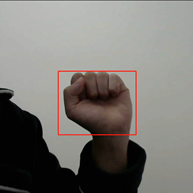}
        \caption{Close}
        \end{subfigure}
        \begin{subfigure}[b]{.3\linewidth}
        \centering\includegraphics[width=.5\linewidth]{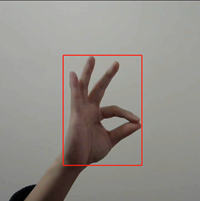}
        \caption{Pinch Close}
        \end{subfigure}
        \begin{subfigure}[b]{.3\linewidth}
        \centering\includegraphics[width=.5\linewidth]{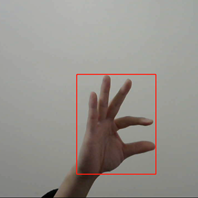}
        \caption{Pinch Open}
        \end{subfigure}
        \begin{subfigure}[b]{.3\linewidth}
        \centering\includegraphics[width=.5\linewidth]{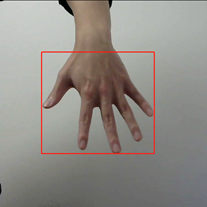}
        \caption{Flip}
        \end{subfigure}
    \caption{TAS Test hand gestures demonstration}
    \label{fig:my_label4}
\end{figure}
Quantifying the blurriness of the images is essential for us to classify and pre-process the training data. Pech-Pacheco et al. proposed a method to calculate the blurriness of the images by calculating the standard deviation of a convolution operation after a Laplace mask \cite{pech2000diatom}.Figure 3 shows the score after calculation, if an image has a high variance, meaning there are many edges in the picture, which is commonly seen in a normal, accurately focused picture. On the other hand, if the picture has a small variance, meaning that there are fewer edges in the picture, which is often seen in a blurred image. We have calculated the blurriness of 20,000 images and classified the quality of the images into three categories, they are clear (Fuzzy > 50), blurred (50 < Fuzzy < 10), and totally burred (Fuzzy < 10) respectively. Table 3 shows how many clear images and blurred images are distributed in each dataset.
\begin{figure}[h]
    \centering
        \begin{subfigure}[b]{.3\linewidth}
        \centering\includegraphics[width=.5\linewidth]{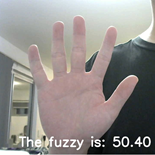}
        \caption{Clear Image}
        \end{subfigure}
        \begin{subfigure}[b]{.3\linewidth}
        \centering\includegraphics[width=.5\linewidth]{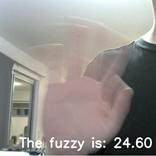}
        \caption{Blurred Image}
        \end{subfigure}
        \begin{subfigure}[b]{.3\linewidth}
        \centering\includegraphics[width=.5\linewidth]{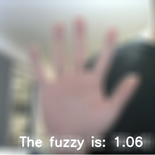}
        \caption{Totally Blurred}
        \end{subfigure}
    \caption{Image blurriness calculation}
    \label{fig:my_label5}
\end{figure}

 \begin{table}[ht]
    \centering
    {\begin{tabular}{p{0.08\textwidth} p{0.08\textwidth} p{0.08\textwidth} p{0.08\textwidth} p{0.12\textwidth} }
    \hline
        Dataset & Total & Clear & Blurred & Totally burred\\
    \hline
        Dataset 1 & 5,376 & 5,376 & 0 & 0  \\
        Dataset 2  & 6,450 & 5,376 & 859 & 215  \\
        
    \hline
    \end{tabular}}
    \caption{Number of images included in the datasets}
    \label{tab:table6}
\end{table}

\section{Methodology}
\subsection{Experiment design}
In our experiment, YOLOV5 is implemented as our object detection model, and we will embed different models or network structures into the object detection model to com-pare the performance. The following network structures will be tested, DarkNet-53 \cite{redmon2018yolov3}, GhostNet \cite{han2020ghostnet}, TinyNet \cite{redmon2018yolov3}, CSPNet-53 \cite{glenn_jocher_2021_4679653},  MobileNetv3-small\cite{howard2019searching},and EfficientNetB1 \cite{tan2019efficientnet}.We will compare the accuracy and speed between different network structures.

\subsection{Experiment Setting}
All the models are embedded in the YOLOV5 detector. In the default settings, we have set the training images as image size 640 x 640 pixels and the training batch size as 32. Then, we have used the following command to train our model from scratch:
python train.py --epochs 50 --data dataset.yaml --cache
We have also implemented several popular data augmentation and training techniques in our experiments. All the hyper-parameters are listed in the appendix Figure. For training techniques, we have set our initial learning rate as 0.01, SGD optimiser is implemented in the training process. Warm-up training is also used in our experiments for all training. For data augmentation, we have implemented translate, scale, flip from left to right, and mosaic augmentation.
\subsection{Quantitative Performance Measures}
The terminologies used to define evaluation metrics are show as below:
\begin{itemize}
    \item Ture Positive (TP): The detection of the object is successful, and the object detected is in the ground-truth bounding box. 
    \item False Positive (FP):  The detection of the specified object is failed with the existence of the object, or false detection is placed in the ground-truth bounding box. 
    \item True Negative (TN): The detection of the object is unsuccessful, however, there is no specified object for detection.
    \item False negative (FN): The detected object is shown on the screen, but the system failed to make a detected ground-truth bounding box.

\end{itemize}
In most of the object detection cases, ground-truth bounding boxes are not perfectly matching the predicted bounding boxes. “IOU” describes the intersection over union (IOU) area, which indicates how well the bounding boxes are localising the objects. Normally, a threshold will be assigned to determine if the detection is successful, if IOU is larger than “t”, it will assume the detection is successful otherwise it will be determined as unsuccessful.
\begin{gather*}
    P(Precision)=\dfrac{TP}{TP+FP}
\end{gather*}
\begin{gather*}
    R(Recall)=\dfrac{TP}{TP+FN}
\end{gather*}
 F1 is also commonly used to comprehensively evaluate the precision and recall of the tested model. F1 will consider both precision and recall in the model evaluation process, F1 score will be 0 if precision or recall equals to 0, which indicates we consider a comprehensive model should have good performance on both precision and recall.
\begin{gather*}
    F1=\dfrac{2*P*R}{P+R}
\end{gather*}
AP (average precision) is regarded as the most common way of measuring the accuracy of the object detection method, which is used to calculate the average precision of each class. mAP refers to the mean value of the average precision, it defines the accuracy of detection for all classes in one database, it is commonly used for measuring the exactness of the detection among all classes in large datasets with a considerable number of classes \cite{padilla2021comparative}. AP can be a useful measurement to deter-mine the accuracy when detecting one specific class 
\begin{gather*}
    AP_{11}=\frac{1}{11}\sum_{R\in\left\{0,0.1,\ldots,0.9,1\right\}}{P_{interp}\left(R\right)}
\end{gather*}
mAP is widely adopted for calculating the average AP for all the classes, for example, YOLO and SSD \cite{liu2016ssd}\cite{r3}. The following equations shows how AP is calculated, AP11 means to split the precision and recall graph into even 11 parts and calculate their average precision accordingly.
\begin{gather*}
    mAP=\frac{1}{N}\sum_{i=1}^{N}{AP_i}
\end{gather*}
In our experiments, we will implement the following evaluation metrics, they are parameters, GFLOPs, weight size, image process speed, mAP@0.5:0.95. Parameters refer to the sum of all weights and biases. In traditional deep learning frameworks, parameter often determines the expressiveness of the model, the more parameters a model has, the more expressiveness it has. GFLOPs are used to evaluate the computational cost in our cases, the more GFLOPs a model has, the greater the cost of the computer to run the model. Weight size is a good indicator to show how big is the model, normally lightweight network structures will have smaller weight sizes. The unit of the weight size in megabyte (MB). Image process speed describes the time spent for the model to process an image with a size 640x640 pixel.mAP@0.5:0.95 is a standard metric from the MS COCO challenge (Lin et al. 2014) and denotes the mean mAP at different IoU thresholds (from 0.5 to 0.95 in steps of 0.05). 
\begin{gather*}
mAP@0.5:0.95=(mAP@0.5+mAP@0.55+\ldots+mAP@0.95)/10
\end{gather*}

\section{Results}
\subsection{Training Results}
\begin{figure}[h]
    \centering
    \includegraphics[scale=1.0]{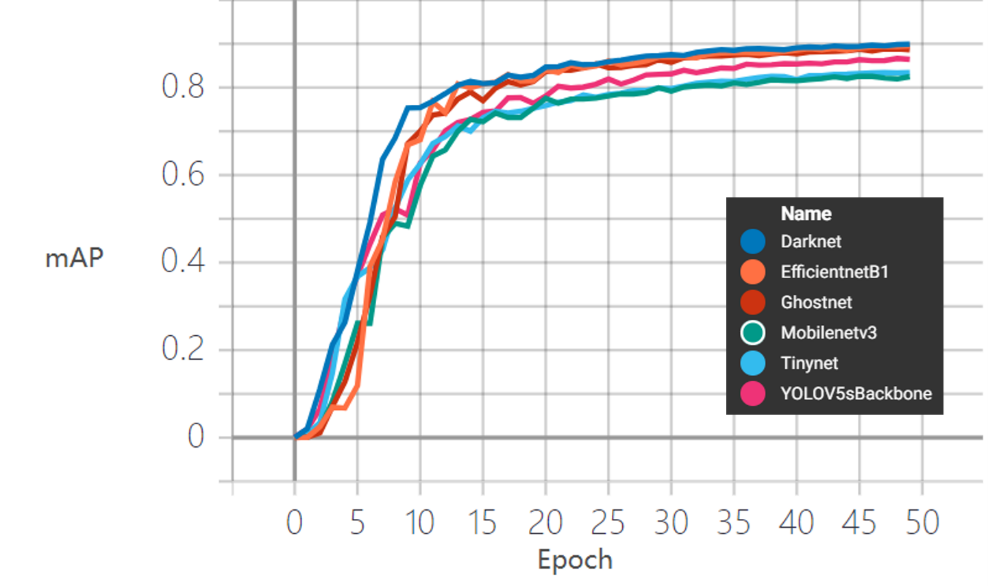}
    \caption{mAP.5:.95 comparison between different network structure on validation Dataset}
    \label{fig:system_overview}
\end{figure}
Figure 4 shows how different network structure performs on validation dataset, it can be found that the map is peaked around 40 epochs at all models. This indicates there is no over-fitting issue with all network structures on our datasets.According to Fig. 3, the mAP values of different models converges steadily, however, it can be found that smaller networks such as MobileNet and TinyNet converge slower than complex networks such as DarkNet and EfficientNet.

\subsection{Comparison between different network structures}
    \begin{table}[ht]
    \centering
    \resizebox{\textwidth}{!}{\begin{tabular}{p{0.22\textwidth} p{0.12\textwidth} p{0.12\textwidth} p{0.10\textwidth} p{0.12\textwidth} p{0.12\textwidth} p{0.12\textwidth}}
    \hline
    Network structure & Parameters & GFLOPs & Weights Size & Process speed per image & mAP@0.5:0.95 \\
    \hline
    GhostNet          & 4.17M      & 9.2    & 7.8MB        & 6.7ms                   & 0.735        \\
    CSPNet53          & 7.27M      & 16.9   & 14.00MB      & 7.3ms                   & 0.753        \\
    MobileNetv3-Small & 3.55M      & 6.3    & 4.21MB       & 3.7ms                   & 0.701        \\
    TinyNet           & 2.18M      & 3.3    & 5.85MB       & 3.1ms                   & 0.703        \\
    Darknet-53        & 6.99M      & 19.0   & 14.20MB      & 7.5ms                   & 0.755        \\
    EfficientNet-B1   & 9.98M      & 6.7    & 19.40MB      & 7.7ms                   & 0.757        \\
    CSPNet53-P6       & 12.36M     & 16.7   & 25.1MB       & 10.5ms                  & 0.776        \\
    Ours              & 12.54M     & 17.0   & 25.5MB       & 10.7ms                  & 0.782      \\
        
    \hline
    \end{tabular}}
    \caption{Backbone comparison on testing dataset (Image Tested on NVIDIA GTX 3090)}
    \label{tab:table7}
\end{table}
 The average speed and mAP were displayed in Table 4. When different network structures encounter data they have not seen before, they do not perform as we expected. Although CSPNet is inferior to Darknet and Efficientnet in the validation dataset, CSPNet outperforms them in the testing dataset. Another observation is that Darknet-53 has a smaller number of parameters than CSPNet whereas its GFLOPs is higher than CSPNet, which means Darknet may require more computing resources to train and test. Moreover, the image process speed of lightweight network structures is 2 times faster than the other network structures whereas their mAPs are significantly less.
\subsection{How transformer block improves performance}
\begin{figure}[h]
    \centering
    \includegraphics[scale=1.0]{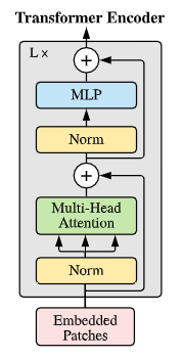}
    \caption{Transformer Block}
    \label{fig:system_overview}
\end{figure}
The author of Vision Transformer has proposed a skip-connection-based network block, transformer encoder as shown in Fig. 6. The transformer encoder network is consisted by multi-layer perception, two norm layers and one multi head attention. It enables the neural network to extract global information from image features \cite{dosovitskiy2020image}. 
    \begin{table}[ht]
    \centering
    \begin{tabular}{p{0.24\textwidth} p{0.15\textwidth} p{0.10\textwidth} p{0.15\textwidth} p{0.15\textwidth}}
    \hline
        Backbone & Process speed per image & mAP-All & mAP Pinch-Open & mAP Pinch-Close\\
    \hline
        YOLOV5s & 8.6ms & 0.753 & 0.541 & 0.741 \\
        V5s+Transformer & 10.5ms & 0.765 & 0.641 & 0.810\\
        YOLOv5-P6 & \textbf{7.4ms} & 0.776 & 0.627 & \textbf{0.827}\\
        P6+Transformer(Ours) & 10.7ms & \textbf{0.782} & \textbf{0.685} & 0.820 \\
    \hline
    \end{tabular}
    \caption{Backbone comparison between with and without transformer layer (Image Tested on NVIDIA GTX 3090)}
    \label{tab:table8}
\end{table}
Table 4 shows how transformer block helps us to improve the performance of distinguishing similar gestures. Although applying transformer block to CSPNet will result an increase in detection speed, the increase is not significant. Even with a detection speed of 11ms per image, the model is still able to achieve real-time detection. From Table 4, we can see that adding the transformer block not only increases the overall gestures detection performance, but also significantly increases the model's detection efficiency for ‘Pinch Open’ pose, which reducing the probability of the model dis-classifying similar gestures. Moreover, by comparing the mAP between the two gestures, we can see that the model with the transformer block has a smaller mAP difference, which also means that the model is more balanced in the classification tasks. Therefore, we have proven that transformer block increases the classification accuracy of similar poses.
\subsection{How different network structure performed with noisy data?}
\begin{table}[ht]
    \centering
    {\begin{tabular}{p{0.18\textwidth} p{0.10\textwidth} p{0.10\textwidth} p{0.10\textwidth} p{0.12\textwidth} p{0.12\textwidth}}
    \hline
        Network structure & mAP without noise & mAP with noise & mAP dropped & Process speed per image \\
    \hline
        CSPNet            & 0.753                      & 0.745                   & 0.008       & 6.7ms                   \\
        GhostNet          & 0.735                      & 0.726                   & 0.009       & 7.3ms                   \\
        MobileNetv3       & 0.701                      & 0.694                   & 0.007       & 3.7ms                   \\
        Darknet-53        & 0.755                      & 0.747                   & 0.008       & 3.1ms                   \\
        EfficientNet-B1   & 0.757                      & 0.745                   & 0.012       & 7.5ms                   \\
        TinyNet           & 0.703                      & 0.701                   & 0.002       & 7.7ms                   \\
        CSPNet-P6         & 0.776                      & 0.769                   & 0.007       & 10.5ms                  \\
        Ours              & 0.782                      & 0.771                   & 0.011       & 10.7ms       \\
        
    \hline
    \end{tabular}}
    \caption{Network performance with noisy data (Image Tested on NVIDIA GTX 3090)}
    \label{tab:table9}
\end{table}
With more data provided (normal data and noisy data), we are expecting the performance will increase, however, the performances of all network structures were dropped. Table 6 shows how different network structures performed with noisy data. Overall, we can find that complex network structures will have higher drop in mAP. There are two possible reasons, on one hand, because their mAP is already higher than simple backbones, on the other hand, complex models usually have deeper neural networks and strong representative ability, So the characteristics of noisy data will be amplified in the representative process. 
\subsection{Attention Layer}
Squeeze-and-Excitation (SE) block enhanced ability of models to learn correlations between channels. SEBlock was implemented in ResNet and improved the performance of ResNet. Although SENet introduces more operations, the performance degradation is still within acceptable limits and the loss of SENet is not very significant in terms of GFLOPs, number of parameters and runtime experiments. Fig. 7 shows the architecture of the SE Block. \cite{hu2018squeeze}.

\begin{figure}[h]
    \centering
    \includegraphics[scale=0.5]{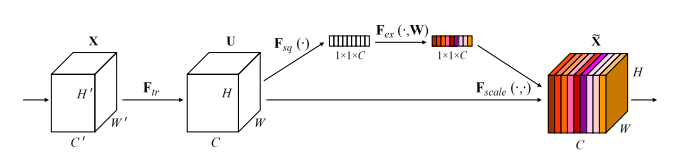}
    \caption{SE Block demonstration}
    \label{fig:system_overview}
\end{figure}
\begin{table}[ht]
    \centering
    {\begin{tabular}{p{0.22\textwidth} p{0.15\textwidth} p{0.15\textwidth} p{0.10\textwidth} p{0.12\textwidth} p{0.12\textwidth}}
    \hline
        Backbone & Precision & Recall & mAP 0.5 & mAP 0.5:0.95 \\
    \hline
        YOLOV5s  & 0.915 & 0.901 & 0.885 & 0.745\\
        YOLOV5s-P6 & 0.915 & 0.902 & 0.887 & 0.761\\
        P6-Transformer & 0.921 &  0.9&0.891 &0.759\\
        P6-SELayer & \textbf{0.927} & 0.914 & 0.903 & 0.765\\
        Trans+SE(Ours) & 0.926 &\textbf{0.923} &\textbf{0.909}&\textbf{0.771}\\
    \hline
    \end{tabular}}
    \caption{Backbone performance with noisy data (Image Tested on NVIDIA GTX 3090)}
    \label{tab:table10}
\end{table}
It has been demonstrated that although the transformer layer enables the network to obtain global information, the quality of the image does not allow the neural network to perform precise feature extraction after transformer layer. Therefore, the introduction of the SE-Block in networks has increased the connection between each layer and thus increased the overall performance.
\section{Conclusions and future work}
\subsection{Summary}
In this paper, we have compared the performance of different neural network structures for hand gesture classification tasks. CSPNet-53 is inspired by the FPN network structure, this type of multi-scale prediction approach is shown to have advantages for hand gesture detection. CSPNet53-P6 network adds a new size of detection head to the CSPNet-53, from 3 output layers to 4 output layers, thus further increasing the accuracy of the model detection. During our experiments, we found that multi-scale prediction is not only useful for improving overall performance, but also for improving the ability of neural networks to distinguish between similar gestures. Our network is inspired by CSPNet53-p6, we have embedded two types of attention mechanism modules, transformer layer and SE-Layer. The trans-former layer enables us to obtain global information and SE-Layer increased the weight of effective features and decreases the influence on ineffective features. Experiments show that adding an attention mechanism further increases the network's ability to discriminate between similar gestures. 
\subsection{Limitation}
While we are discussing the effect of different network structures on the performance of hand gesture detection, we have made many attempts to improve the accuracy of the model and have succeeded in improving the classification accuracy for similar gestures. Essentially, we are sacrificing a portion of the speed of detection to improve the performance accuracy. Moreover, the implementation of the transformer block may require additional computing resources in the training process. 
\subsection{Future Work}
In the ResNet paper, He et al. demonstrated that stacking the depth of the network can improve the expressiveness and performance of the model, but if we keep increasing the depth of the network, it can lead to overfitting the model. He et al. solved this problem from the idea of identity, also known as skip connection, which allows the network to not update parameters if there is nothing learned \cite{he2016deep}. This approach allowed AI to surpass humans in image classification for the first time. We are thinking about the possibility of applying the concept of identity to noisy data immunity so that neural networks can also choose to skip connections when facing noisy data. This can be used to reduce the impact of noisy data on the whole model.
\bibliographystyle{unsrt}  



\bibliography{references}
\appendix
\clearpage
\section{Appendix}
\begin{figure}[h]
    \centering
    \includegraphics[scale=0.6]{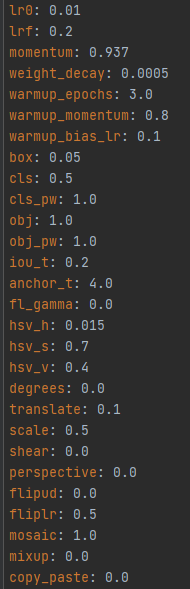}
    \caption{Experiment Hyper-Parameters}
    \label{fig:system_overview}
\end{figure}

\end{document}